# Absolutist AI

Mitchell Barrington

Center for AI Safety
University of Michigan
University of Southern California

## Abstract

This paper argues that training AI systems with *absolute constraints*—which forbid certain acts irrespective of the amount of value they might produce—may make considerable progress on many AI safety problems in principle. First, it provides a guardrail for avoiding the very worst outcomes of misalignment: An AI attempting to commit mass murder *might* have correctly deduced that doing so maximizes expected value, but more likely, the system is severely misaligned. Second, it could prevent AIs from causing catastrophes for the sake of very valuable consequences, such as replacing humans with a much larger number of beings living at a higher welfare level. Third, it makes systems more *corrigible,* allowing creators to make corrective interventions in them, such as altering their objective functions or shutting them down. And fourth, it helps systems explore their environment more safely by prohibiting them from exploring especially dangerous acts. I offer a decision-theoretic formalization of an absolute constraints, improving on existing models in the literature, and use this model to prove some results about the training and behavior of absolutist AIs. I conclude by showing that, although absolutist AIs will not maximize expected value, they will not be susceptible to behave irrationally, and they will not (contra *coherence arguments*) face environmental pressure to become expected-value maximizers.

# 1. Introduction

Advanced AI systems are expected to be dangerous because of the opacity of their goals: We may know that they will effectively pursue their goals but fail to know what those goals are. As a result, it is important to ensure that even if an advanced AI's goals differ from our own, we can minimize the amount of harm they would cause. One way to achieve this result is to ensure that, no matter which goals the system has, they are not the kind of decision maker that would do *anything* to achieve those goals: There are some means to their ends that they would refuse to take. This model of decision-making—where there are some acts that are absolutely prohibited, irrespective of the value they might produce—is familiar in the literature on normative ethics. It is most famously a feature of *absolutist* deontological theories, which posit a set of duty considerations (e.g., the duty to avoid killing people) that take absolute precedence over consequentialist considerations (e.g., the duty to cure headaches) such that no number of cured headaches could justify a killing. This paper argues that there are substantial safety reasons to program absolute constraints into advanced AI systems (§2), offers a novel decision-theoretic model of absolutism (§3), and discusses the implications for the training and behavior of the absolutist AI systems (§4).

# 2. Safety Advantages

A wide range of reasons might support training absolute constraints into AI systems. The most obvious is that doing so might be required to have morally aligned AIs: If moral absolutism or some consequentialist theory positing *value superiority* is true, then designing AIs to respond appropriately to moral considerations might entail training them to treat some considerations as taking absolute precedence over others.[1] This paper puts aside these considerations, instead arguing that there are considerable safety advantages to training absolute constraints into AI systems. This section discusses four: alignment, superhuman moral status, corrigibility, and safe deployment. In each case, even if absolute constraints do not solve the problem in its entirety, they are compatible with a wide range of other possible solutions; they appear to at least be a useful weapon in our arsenal.

---

[1] Even if we are quite confident that it is untrue as a normative ethical theory, considerations of moral uncertainty might favor implementing *some* absolute constraints—where the duty violation would be particularly severe and the net benefit on a consequentialist calculus would be relatively slight.

## 2.1 Alignment Failsafe

An AI system is *aligned* when it behaves how we want it to. Unfortunately, creating aligned AI systems is challenging: It is incredibly difficult to give an exclusive and exhaustive list of everything we care about. And even if we could, designing an objective function that will train the system to be sensitive to everything we care about is not straightforward. Absolutist constraints do not solve the problem in its entirety. However, they might offer an effective safeguard against the worst outcomes of misalignment.

Consider some act *m* and AI agent Aidan who has *m* among his option set. Consider that there is some *m* such that we are very confident that we do not want Aidan to *m* on the grounds of some features of the act. For instance, if *m* involves killing millions of innocent people, our credence that *we want Aidan to m* should be extremely low; as a result, our credence that *Aidan is misaligned* should be much greater than our credence that *we want Aidan to m.* When we discover that Aidan wants to m, we are conditionalizing on the disjunction of these two possibilities: Either we want Aidan to m, or Aidan is misaligned (concerning whether to *m*). Since our prior that *we want Aidan to m* is much lower than our prior that *Aidan is misaligned,* and conditionalizing on *Aidan wants to m* means conditionalizing on their disjunction, we should be more confident that an Aidan who wants to *m* is misaligned. Applied to this case, if Aidan wants to kill millions of innocents, we should conclude that Aidan is misaligned, not that we want him to perform this act. So, we should expect a (sufficiently well-articulated) prohibition on *m* to improve Aidan's behavior.[2]

---

[2] The *well-articulated* stipulation is to avoid having the system find loopholes in the constraint that are just as bad or worse as the prohibited behavior. For instance, we would not want to prohibit an AI from killing only to incentivize making the lives of humans so miserable that they commit suicide.

In addition to safety problems, we can apply this solution to issues in AI fairness. For instance, we might want an AI system to avoid discriminating based on some group categories, such as race, class, or gender identity. While strategies like carefully curating training data will be important for securing this result, we cannot eliminate the possibility that a system will nevertheless determine that it can maximize reward by considering these factors. Implementing an absolutist prohibition on discriminatory policies (perhaps articulated as acts whose expected value is sensitive to the truth value of propositions concerning the individual's group membership) might then be a safeguard for ensuring the system's fairness.

## 2.2 Catastrophic Value Maximization

Future technological advances may make it possible to create beings who convert resources into wellbeing much more efficiently than humans (Shulman and Bostrom 2021). For instance, suppose the resources required to make one hundred humans happy could instead be used to make one million *super-beneficiaries* ten times happier. AI systems trained to optimize wellbeing might see replacing humans with these super-beneficiaries as a substantial improvement. But—for many moral reasons that are evident to everyone except the most devout utilitarians—a small number of engineers do not have the right to create technology directed at destroying the lives and communities of people around the world. But in addition to these moral reasons, there might also be procedural ones: We do not allow elected officials to be unconstrained expected-value maximizers; plausibly, similar procedural considerations should constrain those summoning technology that will affect all of humanity.

We might be able to avert this kind of catastrophe by programming the AGI with a faulty axiology: training it to act as though some collection of humans is, in fact, more valuable than any collection of super-beneficiaries. But we should be careful about deliberately feeding false propositions to a sufficiently advanced learner. Let $f$ be the false proposition we want the AGI to believe. The solution here consists of selecting some $f$ such that the target action (in this case, not causing the extinction of humanity) is rational on $f$ but not on ¬*f*. The first problem is that this solution might not be robust in the face of new evidence: Since *f* is false, the AGI may acquire strong evidence that *f* is false. And since there will be pressures to stamp out contradictions—a system cannot maximize reward by behaving as though *f* is both true and false—it may discard its belief that *f*.

Suppose we patch the problem by designing the system with an extremely high credence in *f*. That way, it will not update very strongly on evidence against *f* and so retain its belief in the face of strong evidence. As a result of *f* being retained, the target action is rational. But the effects of this solution will be impossible to quarantine to this one action. Any proposition whose probability correlates with *f*'s will be affected by the system's credence in *f*; the false belief will infect the belief system. This infection causes more actions to be rational on *f* that would not have been rational on ¬*f*. For instance, conditional on *f* (that some collection of humans is more valuable than any collection of super-beneficiaries), it might be rational for the system to believe *d*: that humans are particularly valuable due to their DNA structure.—Why else would they be more valuable than non-human creatures with identical phenomenal experiences?—And conditional on *d*, it might be rational to replace humans with a large number of non-sentient creatures with human DNA. Fundamentally, the problem is that a high-intelligence AI system will competently deduce the consequences of its beliefs, false propositions imply false propositions, and agents acting on false beliefs are disposed to produce undesirable outcomes.

Absolutist AI, on the other hand, will perform the target action not because of a false axiological belief but because of a duty to do so. This system might well believe that, for any population of humans, there is some population of super-beneficiaries that would be more valuable. However, it would not convert this value into action. The only remaining problem is one of alignment: We must ensure that the prohibition we feed the system covers all the actions we want to prohibit.

## 2.3 Safe Deployment

Reinforcement learning systems learn by exploration and exploitation: They *explore* the environment by gathering information about how much reward they get from performing different acts; they *exploit* the environment by using this information to maximize reward.

But in order to explore, systems must perform non-optimal acts. The problem is that some non-optimal acts would have disastrous consequences, so we do not want the system to try them even once (Amodei et al. 2016).

The most straightforward solution to safe exploration involves training systems in a sandbox, where they can learn about their objective function in a safe, isolated environment. For instance, we might seek to train a self-driving car in a simulation where it can learn to avoid hitting pedestrians. But there are at least three problems with this approach. The first problem concerns exploitation, while the second two concern exploration. There will inevitably be a *simulation gap:* The simulation will never perfectly model all the relevant features of the real world, so some policies that succeed in the simulation domain will fail to generalize (and, of course, some solutions in the real world will be unable to be discovered in the simulation (e.g., Bird and Layzell 2002)). So, no matter how well systems are trained in the sandbox, there will be some risk that they will perform extremely harmful actions after deployment. Second, we do not want our systems to focus solely on exploitation after deployment: We want them to continue to learn, which means exploring and sometimes performing non-optimal acts. However, when systems explore after deployment, they perform non-optimal acts, and some non-optimal acts would be catastrophic. The puzzle is to make systems learn without attempting to learn from catastrophes. And third, even if we can get our agent to explore safely after deployment, we will be less in control of the data it learns from, possibly allowing it to learn harmful behaviors. The most obvious example of this is chatbots learning to be racist from their interactions with humans, such as Microsoft's *Tay* and Meta's *Blenderbot 3.*

Training AIs to be absolutists offers some protection against all three concerns. First, an absolutist constraint against some extremely harmful acts would provide an additional guardrail against a catastrophic outcome of a simulation gap: If a system learns some extremely harmful behavior due to a simulation gap, a prohibition on that behavior may prevent the system from causing a catastrophe. Second, a system that can explore the real world would be much safer with an absolute prohibition on exploring extremely harmful acts. This system would benefit from exploration with very little cost: The only acts they would be prevented from learning about have an extremely low probability of being optimal, so there is little value in learning about them. And third, AIs with out-of-the-box constraints against certain harmful behaviors will be robust against learning that harmful behavior from bad data.

## 2.4 Corrigibility

A *corrigible* AI system allows its creators to make corrective interventions, such as altering its objective function or shutting it down. It is an important component of safe AIs and a particularly difficult property to implant in a system. Corrigibility research has reached something of an impasse: Many believe that advanced AI agents will adopt a value function and proceed to maximize expected value (Omohundro 2008: 10; 2008; Armstrong 2009; Yudkowsky 2015, 2016, 2019; Shah 2018; Grace 2021, 2021; Millidge 2023). So, standard approaches to corrigibility involve describing a decision procedure according to which, conditional on a "shutdown button" being pressed, shutting down maximizes expected value, and conditional on the button not being pressed, not shutting down maximizes expected value—but the system has no preference over *whether* the button is pressed (and so will not either press or disable it). Unfortunately, balancing this indifference is nearimpossible, and stipulating it is problematic (e.g., Soares, Armstrong, and Yudkowsky 2015; Armstrong and O'Rourke 2018).

A more promising solution emerges when we give up the assumption that AIs must act as expected-value maximizers. Absolutist AIs do not face the balancing problem: They do not need to be indifferent between the button being pressed and not being pressed. They prefer that the button is not pressed, but they recognize a duty to avoid interfering with it being pressed and another duty to comply. This solution reduces corrigibility to the more general task of alignment: The system will be motivated to find loopholes in their duties, so we must ensure the duties are specified precisely (outer alignment) and are trained into the system adequately (inner alignment).

## 3. Decision-theoretic Absolutism

I will model absolutism as *expected lexicographic value maximization*—with some modifications. On this approach, the choiceworthiness of an act is treated as a *vector quantity*—that is, as having multiple dimensions—where the *constraint* component is infinitely weightier than the *value* component. So, agents will maximize expected *deontic weight* (the degree to which, in some outcome, they would uphold their duties) and then break ties with expected *value.* An act's expected deontic weight is intended to measure the weight of the duty-based reasons favoring that act. For instance, an outcome in which an agent kills two people presumably has twice the negative deontic weight as an act in which they kill one person (stipulating that the deontic weight of killings is linear).

Unfortunately, absolutist moral theories famously have trouble with risk: That one has a duty to avoid killing does not tell us how much *risk* of killing it is permissible to take.[3] Forbidding only those acts that are *certain* to kill would permit AIs to take impermissibly high risks of killing, and forbidding acts with *any* risk of killing would forbid the system from taking small, intuitively permissible risks. So, for such theories to be action-guiding in the real world, we must tell a story about how these constraints are sensitive to risk.

The modified version I propose involves maximizing *rounded* expected lexicographic value. On this view, agents assess the expected deontic weight of acts, then *round* those expectations to the nearest *n,* where *n* is some contextually determined number. That way, acts whose risks of violating a duty are sufficiently small will have their expected deontic weight rounded back to zero, allowing consequentialist considerations to justify the risk. Ties in rounded expected deontic weight are broken by (unrounded) expected value. So, since the expected (negative) deontic weight of driving to the supermarket is sufficiently low, it will be rounded to zero; if the value of driving to the supermarket is greater than the disvalue of the small risk of killing, one is permitted to drive.

It is worth emphasizing that, in general, considerations that make a difference to deontic weight should also make a difference to value: Killing is not merely wrong because it violates a duty (and so has negative deontic weight) but also because it is bad (and so has negative value). So, an act expected to kill *more* people should (*ceteris paribus*) be worse in its expected deontic weight *and* its expected value. As a result, if two acts have different expected deontic weights but are nevertheless *rounded* to the same number, agents will nevertheless prefer, on the basis of expected value, the one with a smaller expected duty violation.

---

[3] Problems of this form are offered by: McKerlie (1986); Ashford (2003: 298); Jackson and Smith (2006, 2016); Colyvan, Cox, and Steele (2010); Huemer (2010); Fried (2012); Sobel (2012); Isaacs (2014); Holm (2016); Tenenbaum (2017); Alexander (2018).

Note a similarity between this *rounding* approach and the *discounting* approach that has been defended in the contexts of decision theory, ethics, and law (e.g., Bernoulli 1738; d'Alembert 1761; Buffon 1777; Condorcet 1785; Borel 1962; Kagan 1989; Jordan 1994: 21718; Aboodi, Borer, and Enoch 2008; Hawley 2008; Haque 2012; Buchak 2013: 73-74; Smith 2014, 2016; Bjorndahl, London, and Zollman 2017; Chalmers 2017; Lazar 2017; Schwitzgebel 2017; Robert 2018; Lee-Stronach 2018; Tarsney 2018; Monton 2019). According to discounting views, sufficiently small probabilities should be rounded down to zero. I have argued against these views elsewhere ([citation withheld]), partly on the grounds that it offers a one-size-fits-all approach with respect to the *severity* of duty violations: When choosing between a small risk of killing one person and the same risk of killing a billion people, we should prefer the former. However, if all small probabilities are rounded to zero, we will be indifferent. My approach leaves small probabilities intact but rounds *expectations.* And since the expected negative deontic weight of killing billions will be far greater than that of killing one, we will prefer to risk killing the one.[4]

Another advantage our view has over discounting views is that our approach gives plausible prescriptions in moral dilemmas: When any choice would result in a duty violation. On the discounting approach, when the probability of violating a duty is too great to be ignored, agents must always take the smallest risk of violation. But this is implausibly extreme: Suppose one is choosing between A and B, and both acts take an impermissibly high risk of killing. The only differences between them are that A's risk is ☐ greater than B's (where ☐ is an arbitrarily small number), and A would cure billions of headaches. Discounting views claim that one should always take the smaller risk at any consequentialist cost. On the rounding view, however, A and B will almost certainly be rounded to the same number, and the billions of headaches cured by A will outweigh the arbitrarily small additional risk. (The only exception to this general rule is when the ☐ additional risk would push A over the critical threshold, forcing A to be rounded to a different value than B. Of course, this feature is desirable in a model of absolutism since if we *always* allowed consequentialist value to justify increased risk, we would simply be maximizing expected value. The implausibility of the discounting approach in moral dilemmas is simply that it *always* prescribes taking the smallest possible risk.)[5]

---

[4] I have developed this theory of absolutism in greater detail and accuracy elsewhere [citation withheld]. The simplified version here is merely a proof of concept sufficient for proving the relevant results in §4.

## 4. Objections

With these decision-theoretic resources, we can prove that absolutists do not maximize expected value. In this section, we see why absolutists cannot be represented as maximizing expected value, then draw out some implications for training absolutist AI. Then, we consider two possible criticisms on the grounds of this failure of representation. First, we consider whether absolutist AIs would be irrational in a way that would prevent them from effectively pursuing our ends. Second, we consider whether absolutists would be pressured by their environment to begin maximizing expected value—making advanced AIs learn to discard their duties.

### 4.1 Absolutists Violate the Sure Thing Principle

Absolutists cannot be described as maximizing expected value because they violate the Sure

---

[5] It is worth noting one result we may want to avoid. Suppose that, instead of performing a single act *a* that would certainly kill someone, we split the risk of killing into a series of *n* acts $a_1...a_n$, each with a 1/*n* probability of killing. We do not want our theory to say that while *a* is forbidden, it is permissible to perform each act in the series $a_1...a_n$ since this theory would violate Weak Agglomeration (Hare 2016: 460). We can avoid this result by ensuring our rounding is sensitive to diachronic considerations, such as how much risk the agent has already taken. For instance, we might make the rounding more fine-grained (thus allowing less risk to be taken) if one has already performed $a_1$ than if they have not.

Thing Principle (and equivalent axioms, e.g., Independence in Von Neumann and Morgenstern (1944)).

> **Sure Thing Principle (STP):** If conditional on E, you ought to A, and conditional on ¬E, you ought to A, then you ought to A (Savage 1954).[6]

Consider a case:

> *Ronnie's Revolver:* Ronnie has an absolutist duty to avoid killing but no duty to save lives. He is aiming a revolver with 100 chambers at Amber; he knows precisely one chamber is loaded. If Chamber 1 is loaded, Pulling would kill her, but ¬Pulling would result in many people dying. If any other chamber is loaded, Pulling would give Amber a small fright, and ¬Pulling would do nothing. (That is to say, there are *only* lives on the line if Chamber 1 is loaded.)

Table 1

|  | Chamber 1 (0.01) | Other Chamber (0.99) |
|---|---|---|
| Pull | Kill | Scare |
| ¬Pull | Lots of Deaths | Nothing |

Stipulating that a 0.01 risk of killing one person is permissible, the expected deontic weight of Pull will be rounded to zero, and Ronnie ought to break the tie with expected value. Then, if the number of deaths in Lots of Deaths is sufficiently great, Ronnie ought to Pull. But notice that, conditional on Chamber 1, Ronnie ought to ¬Pull (since no number of deaths could justify certainly killing), and conditional on Other Chamber, Ronnie also ought to ¬Pull (since it is better to avoid scaring Amber for no reason). And yet unconditionally, he ought to Pull. Ronnie violates STP.

---

[6] Note a complexity: In standard expected value theory, we can articulate STP equivalently in terms of *value* rather than *ought* since what one ought to do (according to EV theory) is always what maximizes expected value. Absolutists also violate the *value articulation* of STP—very straightforwardly—by refusing to violate duties for the sake of sufficiently great amounts of value. Since the behavior of the absolutist supervenes on *ought* but not *value,* the *ought articulation* is more relevant to worries about the rationality of absolutist AIs (discussed in §4.3 and §4.4).

Notice there is a principled explanation of why Ronnie violates STP, though. *If* he were certain that Chamber 1 obtained, then a constraint would prevent him from Pulling, and *if* he were certain that Other Chamber obtained, then there would be no consequentialist incentive to Pulling. But Ronnie's uncertainty falls within the sweet spot of not generating a constraint but keeping the consequentialist benefit in play: He is *not* bound by a duty to avoid Pulling, and there *is* an incentive to Pull. So, he ought to Pull. At least on the surface, this STP violation appears plausible by the absolutist's lights.[7] In §4.3 and §4.4, we will consider some objections to making our absolutists fail to maximize expected value. But first, we will look at what training an absolutist AI system might look like.

---

## 4.2 How to Train Your Absolutist

While we cannot be especially confident of the methods used to train AIs with agential capacities in the future, we can say something about how training absolutist AIs might look today. The current training method that seems most apt to producing agential AI systems is reinforcement learning (RL) (e.g., Mnih et al. 2013; Mnih et al. 2015; Hendrycks et al. 2021): where systems learn by acting, observing a reward signal, and altering their dispositions to maximize reward. Training absolutist AIs with reinforcement learning means rewarding systems for doing what they ought to, not for producing a good state of the world. Absolutists do not have a special theory of value; they have a special theory of converting value into reasons for action. As a result, they sometimes ought not to perform the act that produces the most valuable outcome. When value and reasons for action come apart, reward must track the latter. This feature of absolutism might seem at odds with the spirit of reinforcement learning. But there is no deep reason we cannot use the technique to train systems to uphold absolute constraints.

One way of using reinforcement learning to train an absolutist AI might be to use a *risk-sensitive* reward function that is sensitive not only to the value of the actual outcome of the act but also to how much risk of violating the agent's duties it took. This approach constitutes a substantial departure from orthodox RL training. Perhaps most counterintuitively, it might involve punishing the system for producing good outcomes. For example, consider some painkiller that would alleviate a patient's headache but has a 50% chance of killing them. Ordinary RL agents learn not to prescribe the painkiller by being rewarded when it cures the patient's headache and *strongly* punished when it kills them. Absolutists, on the other hand, learn not to prescribe the painkiller by being punished for prescribing it—*even when it works*—due to its unacceptable level of risk.

---

[7] For a longer discussion of why it is rational for absolutists to violate STP in these cases, see [citation withheld]. Ultimately, given plausible assumptions, the absolutist must choose between STP and transitivity.

Another way we might train absolutist AIs is to train two modules separately. For instance, the system might consist of a maximizer and a vetoer. The maximizer is an ordinary reinforcement learning algorithm that simply tries to maximize reward. However, the maximizer's only way of interacting with the world is through the vetoer, trained to identify unacceptable risks based on their deontic weight. The vetoer then refuses to carry out orders involving unacceptable risk, forcing the maximizer to only suggest acts whose risks are acceptable.[8] (A similar architecture has been used successfully in limited contexts by Hendrycks et al. (2021).) Obviously, there are additional technical challenges present in these training protocols that are not for training non-absolutist AIs. Perhaps these challenges will not exist for future training methods or will be easier to overcome, or perhaps the safety benefits that absolute constraints would provide will make it worth solving these challenges.

### 4.3 Objection 1: Irrationality

According to Expected Value (EV) theory, maximizing expected value is constitutive of instrumental rationality. So, we might think that absolutists—and thus AI systems we train to be absolutists—will behave irrationally and so would not be the kind of agents we would want to pursue our goals. The strongest arguments for EV theory are that the axioms sufficient for representation as an EV maximizer are substantively true: *Value-pump*

---

[8] One complication with these approaches is that, ordinarily, we train for sensitivity to risk via the frequency of outcomes. So, if this is the case for our absolutist AI, the system must learn the probabilities of the relevant outcomes before it can be trained to recognize risk.

*arguments* for each axiom purportedly show that agents who violate it will pursue policies certain to make them worse off by their own lights.[9]

But it is worth getting clear on what exactly these pragmatic arguments show. They show that the violator will perform an act that will certainly bring about a worse outcome than they otherwise could have. But absolutists do not view outcomes as the only standard by which to evaluate acts: That sometimes we ought not to bring about the most valuable outcome—when doing so would violate a duty—is just what it *means* to be an absolutist. So, an agent who refuses to violate their duty for a greater amount of value is not made worse off *by the lights of the absolutist.*

### 4.4 Objection 2: Robustness

One might think that, even if violating STP is not irrational by the lights of absolutism, AI systems will be pushed by their environment to maximize expected value. *Coherence arguments* claim that if an agent is sufficiently advanced, they will see that there are scenarios where they could be value-pumped—in which they would fail to maximize reward—which will incentivize them to conform to the EV axioms (Omohundro 2008: 10; 2008; Armstrong 2009; Yudkowsky 2015, 2016, 2019; Shah 2018; Grace 2021, 2021; Millidge 2023). If these arguments are sound, we may have reason to think that absolutist constraints will not be robust in advanced agents: AGIs will learn to stop being absolutists.

The status of coherence arguments is unclear in general; it is contentious that reinforcement learning agents will learn to maximize EV (see, e.g., Thornley 2023). But in the case of absolutist AI, it is clear that these arguments do not go through. In §4.2, we considered two ways to train an absolutist AI with reinforcement learning; in neither case is it plausible that the system could learn to maximize EV. For the first system, which is trained on a risk-sensitive reward function, it will simply be untrue that in value-pump scenarios, the system will fail to maximize reward: The system's reward signal will punish the system for taking too great a risk of killing, irrespective of the reward, so the system will learn that in forced choices between killing and producing an arbitrarily large quantity of disvalue, killing would *not* maximize reward. In this case, coherence arguments assume that any AI system must be trained on a reward function that is only responsive to the value of outcomes, which is untrue of this training regimen. The second kind of system we considered consisted of two modules: one attempting to maximize reward (Module 1) and another vetoing the acts that take too great a risk of violating a duty (Module 2). Since Module 1 will learn to determine which acts Module 2 will disapprove of, and Module 2 will veto acts on the basis of their deontic weight, the Module 1 will learn to assign no value to acts with sufficiently great deontic weight—anticipating that they will be vetoed.

---

[9] The exception is Completeness: We cannot show that agents who violate Completeness must pursue dominated strategies; at most, we can show they *may* pursue them (Gustafsson 2022: 24-39). To make things complicated, Completeness is also a necessary assumption for the for other valuepump arguments to show that violators will pursue dominated strategies.